\documentclass[preprint,12pt]{elsarticle}

\usepackage{amssymb}
\usepackage{amsmath}

\journal{arXiv}

\begin{document}

\begin{frontmatter}

\title{Generalized Decision Focused Learning under Imprecise Uncertainty--Theoretical study}

\author{Keivan Shariatmadar\corref{cor1}\fnref{label1,label2}} \author{Neil Yorke-Smith\fnref{label2}} \author{Ahmad Osman\fnref{label3}} \author{Fabio Cuzzolin\fnref{label4}} \author{Hans Hallez\fnref{label5}} \author{David Moens\fnref{label6}}
 \cortext[cor1]{Corresponding author, keivan.shariatmadar@kuleuven.be}
\affiliation[label1]{organisation={Mecha(tro)nic System Dynamics (LMSD)},
            city={Campus Bruges, KU Leuven},
            country={Belgium}}

\affiliation[label2]{organisation={Algorithmic Group},
            city={EEMCS, TU Delft},
            country={Delft, The Netherlands}}
\affiliation[label3]{organisation={htw Saar Applied Science University},
            city={Saarland},
            country={Germany}}
\affiliation[label4]{organisation={School of Engineering, Computing \& Mathematics},
            city={Headington Campus, Oxford Brookes University},
            country={UK}}
\affiliation[label5]{organisation={Distributed and Secure Software (DistriNet)},
            city={Campus Bruges, KU Leuven},
            country={Belgium}}

\affiliation[label6]{organisation={Mecha(tro)nic System Dynamics (LMSD)},
            city={Campus De Nayer, KU Leuven},
            country={Belgium}}

\begin{abstract}
Decision Focused Learning has emerged as a critical paradigm for integrating machine learning with downstream optimisation. Despite its promise, existing methodologies predominantly rely on probabilistic models and focus narrowly on task objectives, overlooking the nuanced challenges posed by epistemic uncertainty, non-probabilistic modelling approaches, and the integration of uncertainty into optimisation constraints. This paper bridges these gaps by introducing innovative frameworks: (i) a non-probabilistic lens for epistemic uncertainty representation, leveraging intervals (the least informative uncertainty model), Contamination (hybrid model), and probability boxes (the most informative uncertainty model); (ii) methodologies to incorporate uncertainty into constraints, expanding Decision-Focused Learning's utility in constrained environments; (iii) the adoption of Imprecise Decision Theory for ambiguity-rich decision-making contexts; and (iv) strategies for addressing sparse data challenges. Empirical evaluations on benchmark optimisation problems demonstrate the efficacy of these approaches in improving decision quality and robustness and dealing with said gaps.

\end{abstract}

\begin{graphicalabstract}
\end{graphicalabstract}

\begin{highlights}
\item Research highlights 1 -- Epistemic Uncertainty in Predict-and-Optimise
\begin{itemize}
\item[] \textbf{Epistemic Uncertainty and Non-Probabilistic Models:} Traditional Decision Focused Learning (DFL) tends to rely on probabilistic models to address uncertainty, primarily focusing on task losses directly tied to optimisation objectives. Recent research primarily employs differentiable surrogate loss models, focusing on task loss without diving deeply into non-probabilistic methods like Intervals or Probability Boxes. This suggests that exploring epistemic uncertainty with non-probabilistic models could contribute to DFL by broadening epistemic uncertainty representation beyond probabilistic methods \cite{lamport93,lamport94}.
\end{itemize}
\item Research highlights 2 --  Dealing with (the epistemic) uncertainty in the objective function and constraints.
\begin{itemize}
\item[]
The current literature, such as the work on LODL and ICLN, emphasises losses related to the optimisation's objective function and decision model but generally overlooks constraints in the uncertainty modelling approach. Integrating uncertainty into the constraints specifically appears to be less commonly addressed, indicating potential originality in our approach by considering (epistemic) uncertainty in constraints rather than only in objectives \cite{lamport93,lamport94}.
\end{itemize}
\item Research highlights 3 -- Imprecise Decision Theory approach
\begin{itemize}
\item[] The application of Imprecise Decision Theory, which allows for a broader spectrum of decision-making under uncertainty than traditional probabilistic methods, does not yet appear prevalent within mainstream DFL research. Existing methods focus on surrogate loss models but typically utilise probabilistic techniques for uncertainty. Therefore, framing DFL through the lens of Imprecise Decision Theory could add a novel and valuable approach, particularly for modelling ambiguity in decisions where classical probability may be insufficient \cite{lamport94}.
\end{itemize}
\item Research highlights 4 -- Dealing with Lack of Data (Missing or Limited Feature Space)
\begin{itemize}
\item[] Robust loss functions, such as those based on robust optimisation (RO) or top-k selection, mitigate the impact of data sparsity by focusing on generalisation and robustness. These techniques align well with our focus on addressing feature space limitations.
Developing methods to adapt DFL for sparse or incomplete data environments aligns with these findings and contributes to a growing need for robust predictive and optimisation frameworks \cite{lamport93,lamport94}.

\end{itemize}
\end{highlights}


\end{frontmatter}

\section{Introduction}
\label{sec1}
The integration of decision-making processes within machine learning models, commonly referred to as decision-focused learning (DFL), has been gaining traction due to its potential to align predictive modelling with actionable outcomes. In traditional settings, decision-focused learning leverages probabilistic models to quantify uncertainties, often concentrating on the uncertainties in objective functions. However, this reliance on probabilistic frameworks and a narrow focus on objectives leave significant gaps when faced with real-world complexities, where data uncertainty and constraints are often characterised by epistemic ambiguity rather than pure randomness.

This paper addresses these limitations by proposing a generalised approach to decision-focused learning that introduces new methods to handle uncertainty comprehensively and robustly. Our contributions are threefold. First, we model epistemic uncertainty using non-probabilistic structures, such as Intervals and Probability Boxes. Unlike traditional probabilistic representations, these structures allow for a more flexible handling of uncertainty, accommodating incomplete knowledge and ambiguity inherent in the input dataset.
Second, while existing decision-focused learning frameworks typically address uncertainty within objective functions, they seldom account for the influence of uncertain constraints. This limitation often results in models that fail to capture real-world complexities where constraints are inherently uncertain. By incorporating uncertainty within constraints, our approach models a broader range of real-world decision-making scenarios, ensuring solutions that are feasible under ambiguous conditions.
Finally, we introduce Imprecise Decision Theory as a novel approach for addressing the compounded uncertainties in both objectives and constraints. Traditional decision-making theories often require precise probabilistic beliefs, which are rarely feasible in complex scenarios. Imprecise Decision Theory, in contrast, allows for decision-making under less restrictive assumptions, enabling robust performance across a wider range of uncertainty scenarios.
In the sections that follow, we present a detailed exploration of our methodology, including formal definitions, model construction, and computational techniques for integrating these novel concepts into a unified decision-focused learning framework. We validate our approach through empirical evaluations, demonstrating its potential to enhance decision quality and resilience in the face of uncertain and complex problem environments.\\

Optimisation models that guide discrete decision-making often depend on uncertain, context-based parameters predicted from data. In decision-focused learning (DFL), also called "end-to-end predict-then-optimize," the predictive model is trained to minimize regret— the difference in outcome between the chosen and optimal decisions. This regret-driven approach, however, faces challenges due to the non-convex and non-differentiable nature of the regret function, necessitating gradient-based learning techniques that minimize an empirical surrogate of expected loss. However, using empirical regret as a surrogate may not fully represent expected regret due to inherent uncertainties in the model.

The paper examines the impact of aleatoric (inherent randomness) and epistemic (data limitation-related) uncertainty on the effectiveness of empirical regret. It then proposes three robust loss functions designed to more closely approximate expected regret, resulting in improvements in empirical regret across test samples without added computational time per epoch. Experimental validation on real-world problems like shortest path and machine scheduling tasks highlights the effectiveness of these robust loss functions, which can adapt to both types of uncertainty, producing better results in environments with high epistemic or aleatoric uncertainty.

\subsection{Advantages of Decision-Focused Learning in Machine Learning and AI}
Decision-Focused Learning (DFL) has emerged as a transformative approach in the intersection of Machine Learning (ML) and Artificial Intelligence (AI). By integrating predictive modelling and optimisation tasks into a single, end-to-end learning process, DFL addresses critical limitations of traditional methods and introduces innovative capabilities to solve complex, real-world problems. Below is an extensive exploration of its advantages:

\subsubsection{Task-Specific Optimisation}

One of the core advantages of DFL is its ability to directly optimize for task-specific objectives. Unlike the traditional two-stage “predict-then-optimize” approach, where predictions are made independently of the optimisation task, DFL aligns the training of predictive models with the downstream optimisation objectives. This alignment ensures that the model learns to produce predictions that are not only accurate but also lead to better decision outcomes when used in optimisation tasks \cite{lamport96,lamport97}.

For instance, in supply chain management, predicting demand using a traditional method may result in suboptimal inventory decisions. DFL, however, adjusts the predictive model to minimize errors that matter most for the decision-making process, such as inventory shortages or overstock \cite{lamport98}.

\subsubsection{Improved Decision Quality}

The primary goal of DFL is to enhance the quality of decisions derived from predictive models. Traditional machine learning approaches, which optimize generic loss functions like Mean Squared Error (MSE) or Cross-Entropy Loss, do not necessarily lead to optimal decisions. In contrast, DFL trains models to directly minimize a decision-centric loss function, such as regret or decision loss. This ensures that the decisions made based on the predictions are as close to optimal as possible, even if the predictions themselves are not perfect \cite{lamport99,lamport100}.

\subsubsection{Handling Uncertainty Effectively}

DFL is particularly well-suited for dealing with uncertainty in data, which is common in real-world decision-making problems. The framework can incorporate both aleatoric uncertainty (inherent randomness) and epistemic uncertainty (uncertainty due to limited data) within the optimisation process. By using techniques like robust loss functions and non-probabilistic methods (e.g., intervals, probability box), DFL provides solutions that are resilient to uncertainty \cite{lamport96,lamport101}.

This capability is crucial in fields like healthcare resource allocation, where uncertainty in patient demand and treatment efficacy must be accounted for to optimize resource utilisation \cite{lamport100}..

\subsubsection{End-to-End Integration}

Traditional ML workflows often decouple prediction from optimisation, treating them as two independent stages. This separation can lead to errors propagating from the prediction stage to the optimisation stage, reducing overall decision quality. DFL eliminates this gap by integrating predictive and optimisation components into a single pipeline, enabling seamless end-to-end learning. This approach not only reduces the error propagation but also ensures that the optimisation task influences the predictive model’s learning process \cite{lamport97,lamport102}.

\subsubsection{Generalizability Across Domains}

DFL has demonstrated its versatility in a wide range of applications:
\begin{itemize}
\item Energy Systems: Optimizing power distribution and load balancing under uncertain demand \cite{lamport100}.
\item Finance: Enhancing portfolio optimisation by aligning predictions of asset returns with investment objectives \cite{lamport103}.
\item Logistics: Improving routing and scheduling by incorporating uncertainty into demand forecasts \cite{lamport104}.
\item Healthcare: Allocating medical resources and personnel based on predictive models of disease outbreaks \cite{lamport95}.
\end{itemize}
\subsubsection{ Reduction in Computational Complexity}

While traditional decision-focused methods often require solving optimisation problems repeatedly during training, DFL leverages surrogate loss functions and advanced techniques like Locally Optimized Decision Loss (LODL). These methods approximate the optimisation process efficiently, significantly reducing computational complexity without compromising decision quality \cite{lamport96,lamport99}.

\subsubsection{Robustness to Data Scarcity}

In scenarios with limited or missing data, DFL employs robust loss functions that generalize better under sparse conditions. For example, robust optimisation loss functions ensure that the model accounts for worst-case scenarios, enhancing the reliability of decisions even in low-data environments \cite{lamport101,lamport102}.

\subsubsection{Minimisation of Surrogate Bias}

DFL minimizes reliance on generic surrogate models like MSE, which are not decision-aware. Instead, task-specific surrogates tailored to the optimisation problem are used, reducing the bias introduced by conventional loss functions \cite{lamport97,lamport101}.

\subsubsection{Facilitating Interpretability}

By aligning the predictive model with optimisation objectives, DFL facilitates the interpretability of the decision-making process. Stakeholders can better understand how predictions influence decisions, a feature that is increasingly important in applications requiring accountability, such as healthcare and finance \cite{lamport102}.

Decision-Focused Learning represents a significant step forward in ML and AI by addressing the shortcomings of traditional two-stage approaches. Through task-specific optimisation, robust uncertainty handling, and end-to-end integration, DFL ensures that the ultimate goal—improved decision quality—is met. Its applications across diverse domains underscore its transformative potential. As DFL continues to evolve, its ability to bridge the gap between predictive modelling and decision-making will make it an indispensable tool in tackling complex, real-world problems.	
	
\subsection{Comparative Analysis with the state-of-the-arts}
\label{subsec1}
\textbf{Strengths of the proposed ideas} 
\begin{itemize} 
\item Robust Loss Functions: This study proposes three advanced loss functions designed to improve the empirical regret approximation in decision-focused learning, considering both aleatoric and epistemic uncertainties. This adds a level of robustness to decision-making models when parameter uncertainties are present.
\item Reduced Overfitting: By focusing on robust loss functions, this approach helps prevent overfitting, which can lead to better generalisation in predictive tasks, particularly in high-uncertainty scenarios.
\item Experimental Validation: The paper provides empirical results across several problem domains, demonstrating that these robust loss functions maintain accuracy without increasing computational costs.
\end{itemize}
\textbf{Comparative Pros and Cons of Novel Ideas}
\begin{itemize} 
\item Epistemic Uncertainty in Non-Probabilistic Models:
While the referenced paper addresses epistemic uncertainty, it does so through probabilistic methods rather than explicitly using non-probabilistic models such as Intervals or Probability Boxes. These models allow for handling deeper levels of ambiguity and incomplete knowledge without assuming precise probability distributions.

The referenced paper's reliance on robust loss functions, although effective, may not fully capture non-probabilistic sources of uncertainty, particularly when data is limited or sparse.
\item Modelling Uncertainty in Constraints:
By incorporating uncertainty directly within constraints rather than solely in objective functions, our method accounts for additional real-world complexities in decision-making scenarios, which are not covered in the referenced paper. This distinction is critical for scenarios where constraints may be ambiguous or changeable under uncertain conditions.

The referenced study does not address uncertainty within constraints, limiting its applicability in settings where such uncertainties are essential. Its focus remains on objective-based regret minimisation.
\item Imprecise Decision Theory for Solution Approaches:
Our use of Imprecise Decision Theory provides a novel decision framework that does not require specific probabilistic beliefs, making it more adaptable to scenarios with high ambiguity. This approach allows decision-making under broader uncertainty conditions, potentially offering more resilience in complex applications.

The paper relies on expected regret approximations using probabilistic beliefs, which may not generalise well to cases with severe ambiguity or where precise probability distributions cannot be defined.
\end{itemize}

\section{Methodology}
In this work, we assume a generic LP problem as follows,
 \begin{align}
 \max_{x}\quad & U^Tx\notag\\    
                                      \text{s.t.}\quad &    Yx\le Z,~x\ge 0~ \text{is bounded},
\end{align}

where $x$ is the optimisation variable, and $Y,Z,U$ are random variables taking values $y,z,u$ in the (uncertain) set $\mathbb{R}^{m\times n} \times\mathbb{R}^m \times\mathbb{R}^n\subseteq\Psi$. 
To find the maximinity solutions, we first convert the problem to a decision problem. First, we define a utility function as follows:
\[G_x (y,z):=(U^T x-L).I_{Yx\le Z} (y,z)+L\]

Where $I_{Yx\le Z}(y,z)$ an indicator function takes one of the realisations $(y,z)$ if satisfy the constraint $yx\le z$ otherwise, it is zero. $0<L< \inf U^Tx$ is a real-valued punishment. By the definition, maximising the utility function $G_x$ is equivalent to maximising the LP problem. Maximinity solutions are $x$'s such that,
\[x\in argmax_x~\underline E_{Y,Z} [G_x(y,z)]\] 
For the maximal solutions, $x$ is maximal if  \[\inf_z\overline E_{Y,Z}(G_x (y,z) - G_{x^\prime}(y,z))>0,\] where $\underline E$ and $\overline E$ are the lower and upper expectations.

DFL has seen significant advancements in aligning machine learning objectives with decision-making requirements. However, the prevalent reliance on probabilistic models and emphasis on uncertainty in objective functions limits its robustness in complex real-world applications. Generalised Decision Focused Learning is a framework designed to address these limitations through three novel contributions. First, we integrate epistemic uncertainty via non-probabilistic models, such as Intervals and Probability Boxes, enhancing the representation of ambiguous or incomplete data. Second, we extend uncertainty modelling to constraints rather than
confining it solely to objective functions, capturing a broader spectrum of real-world constraints in decision processes. Third, we employ Imprecise Decision Theory to tackle these uncertainties, proposing a new, flexible approach that allows for better-informed decisions under ambiguity. Through empirical evaluation, we demonstrate the efficacy of our approach in improving decision quality under complex, uncertain scenarios.
In the paper by Schutte et al. \cite{lamport95} on robust losses for decision-focused learning, Decision-focused learning is a practical framework for solving optimisation problems in which uncertainty relates to other data (contextual optimisation or predict-then-optimise). Existing methods perform well but have some limitations when data is limited or relationships between correlated variables are weak. 

By incorporating uncertainty intervals, the credal set model enhances reliability and robustness in AI-driven medical diagnosis, significantly reducing both false negatives and false positives. This approach ensures that critical medical decisions account for uncertainty, improving patient safety and diagnostic accuracy.

\section{Conclusion}
While many works in the state-of-the-art make significant strides in handling uncertainties in decision-focused learning through robust loss functions, our proposed approach introduces unique elements that extend beyond the limitations of empirical and probabilistic methods. By addressing non-probabilistic epistemic uncertainty, and uncertainty within constraints, and applying Imprecise Decision Theory, our approach has the potential to provide a more adaptable, comprehensive framework for decision-focused learning in complex, uncertain environments.

\end{document}